\documentclass[11pt]{article}

\usepackage{acl}
\setcitestyle{authoryear,round}

\usepackage{times}
\usepackage{latexsym}
\usepackage{amsmath,amssymb,amsfonts}
\usepackage{algorithm}
\usepackage{algorithmic}
\usepackage{graphicx}
\usepackage{wrapfig}
\usepackage{subcaption}
\usepackage{booktabs}
\usepackage{array}
\usepackage{tabularx}
\usepackage{multirow}
\usepackage{makecell}
\usepackage{enumitem}
\usepackage{url}
\usepackage[T1]{fontenc}
\usepackage{cmap}
\usepackage[final]{microtype}
\usepackage[htt]{hyphenat}
\usepackage{xcolor}
\usepackage{placeins}
\usepackage{float}
\usepackage{listings}

\newcolumntype{Y}{>{\centering\arraybackslash}X}

\lstdefinestyle{json}{
  basicstyle=\ttfamily\footnotesize,
  columns=fullflexible,
  breaklines=true,
  breakatwhitespace=false,
  keepspaces=true,
  showstringspaces=false,
  frame=single,
  framerule=0.3pt,
  xleftmargin=0.2em,
  xrightmargin=0.2em,
  aboveskip=0.5em,
  belowskip=0.5em
}
\title{RPMS: Enhancing LLM-Based Embodied Planning through Rule-Augmented Memory Synergy}
\author{
  Zhenhang Yuan, Shenghai Yuan, Lihua Xie \\
  School of Electrical \& Electronic Engineering \\
  Nanyang Technological University, Singapore \\
  \texttt{yuan0142@e.ntu.edu.sg, shyuan@ntu.edu.sg, elhxie@ntu.edu.sg}
}

\begin{document}
\maketitle

\begin{abstract}
LLM agents often fail in closed-world embodied environments because actions must satisfy strict preconditions---such as location, inventory, and container states---and failure feedback is sparse.
We identify two \emph{structurally coupled} failure modes: \textbf{(P1)}~invalid action generation and \textbf{(P2)}~state drift, each amplifying the other in a degenerative cycle.
We present RPMS, a \emph{conflict-managed} architecture that enforces action feasibility via structured rule retrieval, gates memory applicability via a lightweight belief state, and resolves conflicts between the two sources via rules-first arbitration.
On ALFWorld (134 unseen tasks), RPMS achieves 59.7\% single-trial success with Llama~3.1~8B (+23.9~pp over baseline) and 98.5\% with Claude~Sonnet~4.5 (+11.9~pp); of the 8B gain, rule retrieval alone contributes +14.9~pp (statistically significant), making it the dominant factor.
A key finding is that episodic memory is conditionally useful: it harms performance on some task types when used without grounding, but becomes a stable net positive once filtered by current state and constrained by explicit action rules.
Adapting RPMS to ScienceWorld with GPT-4 yields consistent gains across all ablation conditions (avg.\ score 54.0 vs.\ 44.9 for the ReAct baseline), providing transfer evidence that the core mechanisms hold across structurally distinct environments.
\end{abstract}

\section{Introduction}

LLM-driven agents convert natural-language goals into action sequences through iterative environment interaction. However, closed-world embodied environments such as ALFWorld pose distinct challenges: actions must satisfy implicit preconditions (location, inventory, container states), while failure feedback is sparse (e.g., ``Nothing happens''). This gap between linguistic plausibility and environmental validity causes agents to generate actions that \emph{read} correctly but \emph{fail} silently.

\begin{figure}[t]
  \centering
  \vspace{-0.6em}
  \includegraphics[width=\columnwidth]{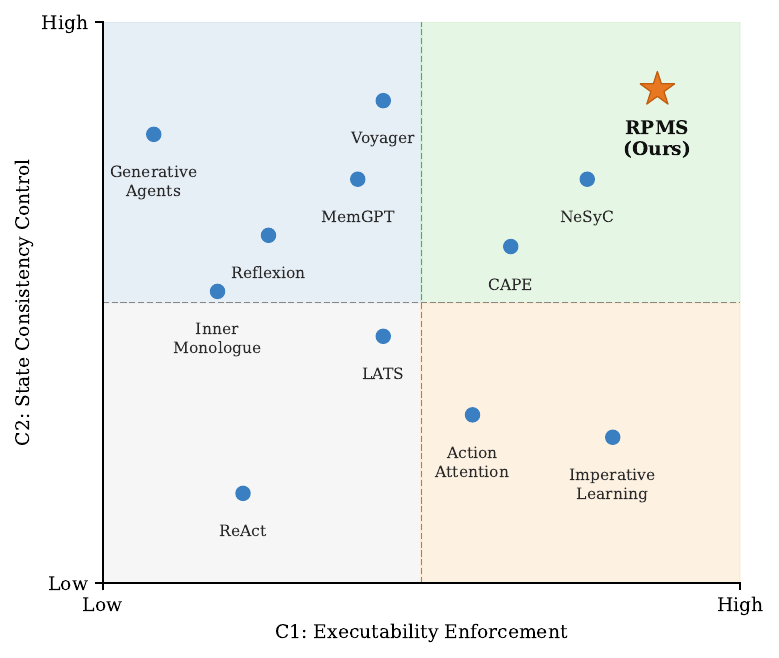}
  \caption{Representative methods positioned along C1 (executability enforcement) and C2 (state consistency control): ReAct~\cite{yao2023react}, Generative Agents~\cite{10.1145/3586183.3606763}, Reflexion~\cite{shinn2023reflexion}, MemGPT~\cite{packer2024memgptllmsoperatingsystems}, Inner Monologue~\cite{huang2022innermonologue}, Voyager~\cite{wang2023voyager}, LATS~\cite{pmlr-v235-zhou24r}, Action Attention~\cite{wu2022tackling}, Imperative Learning~\cite{doi:10.1177/02783649251353181}, CAPE~\cite{raman2024capecorrectiveactionsprecondition}, NeSyC~\cite{choi2025nesyc}. Most prior work addresses one axis; RPMS targets both.}
  \label{fig:teaser}
  \vspace{-0.8em}
\end{figure}

Existing work on embodied decision-making---from imitation learning (BUTLER~\cite{shridhar2020alfworld}) to prompting-based reasoning (ReAct~\cite{yao2023react}) to reflective retry loops (Reflexion~\cite{shinn2023reflexion})---exhibits consistent failure patterns. Two structural constraints explain why. First, actions are restricted to fixed commands (e.g., \texttt{goto}, \texttt{take}, \texttt{heat}), and failures return only weak signals. Second, the environment is partially observable, requiring latent state tracking. These constraints produce two coupled failure modes:
\textbf{(P1)}~\emph{Invalid action generation}---actions violate preconditions that the LLM cannot infer from surface text;
\textbf{(P2)}~\emph{State drift}---the agent's belief diverges from true state, compounding downstream errors.
Critically, P1 and P2 are not independent: state drift causes repeated precondition violations, which produce only uninformative ``Nothing happens'' feedback that further corrupts the belief, creating a degenerative cycle. While prior work addresses P1 or P2 individually through prompting~\cite{yao2023react}, reflection~\cite{shinn2023reflexion}, or memory~\cite{zhang2023rememberer}, explicitly targeting their \emph{joint} interaction as a design objective has not been the central focus of those systems.

We frame these issues as two design challenges: \textbf{(C1) Executability enforcement}---the decision process should expose action preconditions at inference time rather than relying on latent commonsense in model weights; and \textbf{(C2) State consistency control}---retrieved experiences should be checked against the agent's current tracked state before prompt injection, a mechanism not central to prior systems~\cite{yao2023react,shinn2023reflexion,zhao2024expel,zhang2023rememberer}.

These challenges are coupled: injecting action rules without state-consistent memory screening can leave the agent acting on a corrupted belief, while retrieving experience without feasibility grounding introduces advice whose preconditions no longer hold.
We address both with \textbf{RPMS}, a \emph{conflict-managed} architecture in which rule retrieval enforces action feasibility, a lightweight \emph{task-facing belief interface} gates memory applicability, and a rules-first arbitration policy resolves conflicts between the two knowledge sources.

Our contributions are: (i)~a structural analysis of the P1--P2 degenerative cycle, framing knowledge augmentation as the problem of \emph{when knowledge can be safely used at decision time}; (ii)~RPMS, a conflict-managed architecture combining rule retrieval for action feasibility with state-signature compatibility filtering for memory applicability, mediated by a rules-first arbitration policy; and (iii)~a controlled empirical study demonstrating that episodic memory is only reliably beneficial when grounded by explicit action constraints, with transfer evidence on ScienceWorld confirming that the core mechanisms generalise across structurally distinct environments.

\section{Related Work}

\subsection{LLMs for Embodied Decision-Making}

The application of LLMs to embodied task planning has progressed through several paradigms.
Early approaches such as SayCan~\cite{ahn2022saycan} ground high-level language into executable skills via affordance scoring, while BUTLER~\cite{shridhar2020alfworld} learns policies from expert demonstrations in ALFWorld.
Prompting-based agents such as ReAct~\cite{yao2023react} interleave reasoning traces with environment actions, and follow-up work explores richer feedback channels and decomposition into sub-programs or code policies (e.g., Inner Monologue~\cite{huang2022innermonologue}, ProgPrompt~\cite{singh2023progprompt}, Code-as-Policies~\cite{liang2023codeaspolicies}).

For ALFWorld specifically, several lines of work aim to improve long-horizon reliability through supervision, fine-tuning, or iterative refinement.
A3T/ActRe~\cite{yang2024actre} uses autonomous data generation and contrastive self-training to boost success rates.
These approaches often rely on additional training data, parameter updates, or multi-trial refinement, whereas RPMS is \emph{inference-time modular} and targets single-trial reliability through rule grounding plus state-consistent experience retrieval.

\subsection{Reflective and Memory-Augmented Agents}

Reflexion~\cite{shinn2023reflexion} enables agents to learn from verbal self-reflection across multiple trials, storing textual summaries of past failures.
ExpeL~\cite{zhao2024expel} extracts reusable insights from trajectories.
REMEMBERER~\cite{zhang2023rememberer} retrieves relevant past experiences to guide current decisions.
CLIN~\cite{majumder2024clin} builds a continually updated causal abstraction from environment interactions, improving task adaptation across episodes.
In open-ended environments, Voyager~\cite{wang2023voyager} demonstrates lifelong skill acquisition with an external skill library; however, its setting differs from closed-world ALFWorld where action preconditions are strict and failures can be silent.

These approaches demonstrate the value of experiential knowledge, but generally emphasize memory reuse or self-improvement rather than an explicit retrieval-time state-compatibility filter. In closed-world settings, a trajectory from a different initial state can recommend actions whose preconditions are no longer satisfied; RPMS makes this compatibility check explicit (Sec.~\ref{sec:state_filter}).

A key distinction: Reflexion~\cite{shinn2023reflexion} and ExpeL~\cite{zhao2024expel} operate at the \emph{trial level}---retrying the same task across attempts and using each failure to refine the next. RPMS instead operates at \emph{retrieval time within a single trial}: experiences are accumulated offline on seen tasks, then frozen; at test time each recalled entry is screened against the current belief state before entering the prompt, enabling single-trial execution without per-task retry budgets.

\subsection{Knowledge-Augmented Planning}

RAG-based methods inject external knowledge into LLM prompts at inference time~\cite{gao2024ragsurvey}.
In the planning domain, domain-specific rules have been used to constrain action generation~\cite{ahn2022saycan, singh2023progprompt}.
Related lines integrate LLMs with symbolic planning representations (e.g., translating language to PDDL-style plans~\cite{xie2023nl2pddl,guan2023pddlrepair}) to enforce executability, but typically assume access to explicit world states or planners; RPMS instead uses lightweight belief tracking and rule-text grounding without requiring a full symbolic simulator.

Relative to prior knowledge-augmented planning work, RPMS emphasizes two implementation choices central to our setting: a three-tier rule hierarchy (Sec.~\ref{sec:rule_manual}) and an explicit rules-first arbitration policy (Sec.~\ref{sec:arbitration}) for mediating between rule-based and experience-based guidance.
SwiftSage~\cite{lin2023swiftsage} achieves high performance on ScienceWorld through a trained ``SWIFT'' reactive component combined with GPT-4 deliberation; RPMS is complementary in that it achieves consistent improvements through a prompt-only, training-free approach.

\section{Method}
\label{sec:method}

\begin{figure*}[!t]
  \centering
  \includegraphics[width=\textwidth]{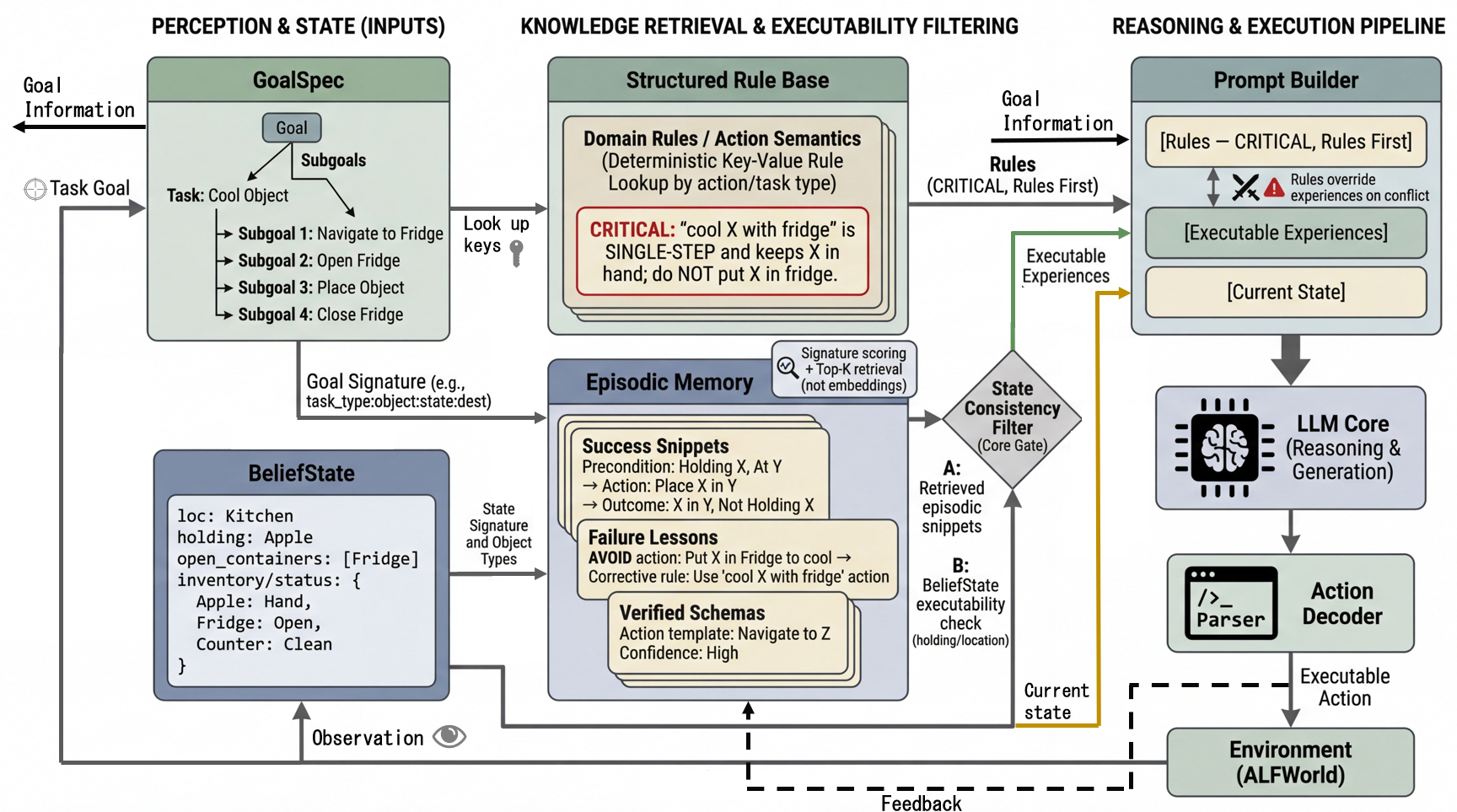}
  \caption{RPMS architecture. Each step: (1) parse observation into \emph{BeliefState} and \emph{GoalSpec}; (2) query Rule Manual and Episodic Memory in parallel; (3) filter experiences by state-signature compatibility; (4) resolve conflicts via Rules-First Arbitration; (5) query LLM with augmented prompt.}
  \label{fig:architecture}
\end{figure*}

\begin{figure}[t]
  \centering
  \includegraphics[width=0.98\columnwidth]{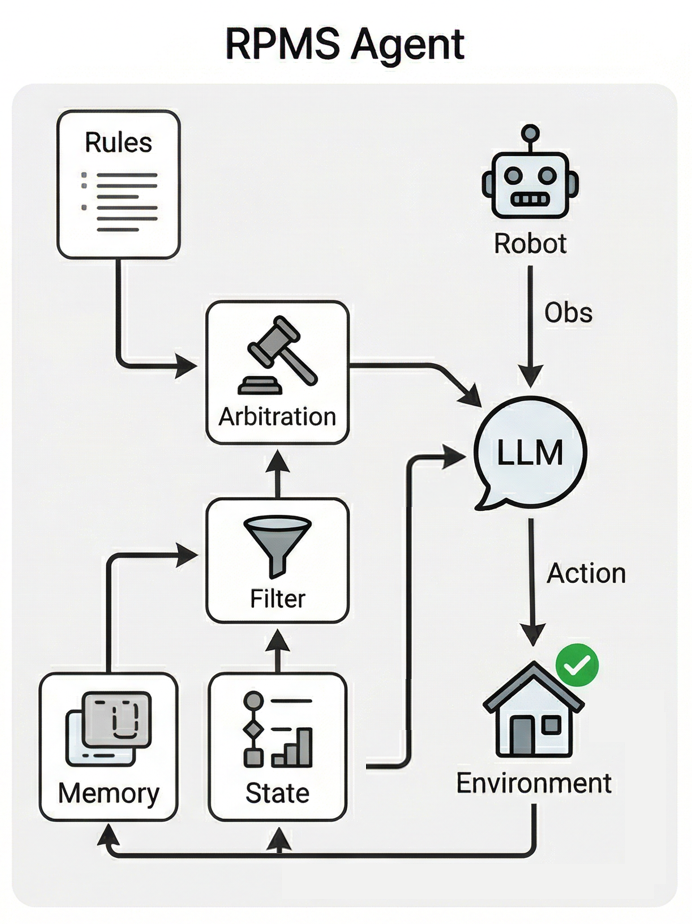}
  \caption{RPMS agent architecture overview, showing rule injection (C1: executability enforcement) and state-consistent memory filtering (C2: state consistency control) as the two core components that augment the LLM decision loop.}
  \label{fig:overview}
\end{figure}

Given an environment observation $o_t$ and a natural-language goal $g$, the agent must select an action $a_t$ from a fixed vocabulary at each step.
RPMS is not a simple combination of rule guidance and episodic memory; it is a conflict-managed architecture for using both under partial observability. Rules $\mathcal{R}$ enforce action feasibility; Episodic Memory $\mathcal{M}$ supplies applicable experience; and a \emph{Rules-First Arbitration} policy resolves cases where the two conflict.
Fig.~\ref{fig:architecture} shows the overall pipeline.

\subsection{Belief State Tracking}
\label{sec:belief_state}

Since the environment emits only textual observations, the agent cannot directly access the underlying world state and must maintain an approximate representation.
Rather than reconstructing a full world model, we maintain a \emph{task-facing belief interface}: only the state variables needed to verify action preconditions and screen memory applicability are tracked.
We define an online \emph{BeliefState} $b_t = \langle \ell_t, h_t, \mathcal{C}_t, \Pi_t \rangle$, where $\ell_t$ is the current location, $h_t \in \mathcal{O} \cup \{\varnothing\}$ the hand state (ALFWorld enforces a single-object grasp constraint), $\mathcal{C}_t$ the open/closed state of openable receptacles, and $\Pi_t$ the last-observed location of each object.
The belief state is updated deterministically:
\begin{equation}
  b_t = f_{\textsc{upd}}(b_{t-1},\; a_{t-1},\; o_t)
  \label{eq:update}
\end{equation}
where $f_{\textsc{upd}}$ applies pattern-matching rules to the observation text (e.g., a successful \texttt{goto}$(l)$ sets $\ell_t \leftarrow l$; \texttt{take}$(o,l)$ sets $h_t \leftarrow o$)---no LLM call is involved.
This design is intentionally minimal: $b_t$ tracks exactly the fields needed for precondition checking, and the same deterministic update principle carries over unchanged to ScienceWorld (where $\ell_t$ becomes a room name and $h_t$ becomes an inventory set).

\paragraph{Sufficiency.}
For the preconditions modeled in our rule manual, $b_t$ provides a practical sufficient summary for action verification.
The following table illustrates the mapping for representative actions:

\begin{center}
\small
\setlength{\tabcolsep}{4pt}
\begin{tabularx}{0.98\columnwidth}{l>{\raggedright\arraybackslash}X}
\toprule
\textbf{Action} & \textbf{Key preconditions on} $(\ell_t, h_t, \mathcal{C}_t)$ \\
\midrule
$\texttt{take}(o,l)$ & $\ell_t {=} l \;\wedge\; h_t {=} \varnothing \;\wedge\; (l \notin \mathcal{L}_{\textit{op}} \vee \mathcal{C}_t(l){=}\texttt{open})$ \\
$\texttt{put}(o,l)$  & $\ell_t {=} l \;\wedge\; h_t {=} o$ \\
$\texttt{heat}(o)$   & $\ell_t {=} \texttt{microwave\,1} \;\wedge\; h_t {=} o$ \\
$\texttt{open}(l)$   & $\ell_t {=} l \;\wedge\; \mathcal{C}_t(l){=}\texttt{closed}$ \\
\bottomrule
\end{tabularx}
\end{center}

\noindent Transformation outcomes (e.g., \textit{heated}, \textit{cooled}) are not tracked in $b_t$ because they appear only in the terminal reward, not as preconditions for subsequent actions.
Static environment properties (action vocabulary, receptacle topology) are time-invariant and encoded in the rule manual rather than in $b_t$.

\paragraph{Signatures.}
For efficient retrieval we compress $b_t$ and goal $g$ into compact keys:
\begin{align}
  \sigma(b_t) &= (\ell_t,\; h_t) \label{eq:state_sig} \\
  \gamma(g) &= (g_{\textit{type}},\; g_{\textit{obj}},\; g_{\textit{dest}}) \label{eq:goal_sig}
\end{align}
The state signature $\sigma$ retains the two components most discriminative for action compatibility; the goal signature $\gamma$ is derived from the natural-language goal via template parsing.

\subsection{Hierarchical Rule Manual}
\label{sec:rule_manual}

The Rule Manual $\mathcal{R}$ organizes action-grounding knowledge into three tiers with decreasing generality (Appendix~\ref{app:rules}):

\begin{itemize}[nosep]
  \item \textbf{Universal rules} $\mathcal{R}_U$: environment-agnostic constraints covering search strategies, failure recovery, and state-management invariants (e.g., complete placement before picking up a new object).
  \item \textbf{Domain rules} $\mathcal{R}_D$: task-type procedures prescribing canonical action sequences and common pitfalls, indexed by goal type so only the relevant subset is injected.
  \item \textbf{Environment rules} $\mathcal{R}_E$: action semantics encoding precondition--effect pairs that deviate from commonsense (e.g., \texttt{heat} and \texttt{cool} are atomic and preserve the hand state---a counter-intuitive constraint that is a frequent source of LLM errors).
\end{itemize}

At each step, $\mathcal{R}_{\text{active}} = \mathcal{R}_U \cup \mathcal{R}_D(g_{\textit{type}}) \cup \mathcal{R}_E$ is assembled and injected into the prompt.
Crucially, the same three-tier structure transfers across environments: adapting to ScienceWorld requires only rewriting the domain and environment tiers, while the universal rules and the injection mechanism remain unchanged.

\subsection{Episodic Memory with State-Consistent Filtering}
\label{sec:state_filter}

The Episodic Memory $\mathcal{M}$ accumulates entries during a prior \emph{learning phase} on seen tasks (Appendix~\ref{app:memory} provides concrete examples).
Each entry $m \in \mathcal{M}$ is associated with a goal signature $\gamma(m)$ and a state signature $\sigma(m)$ recorded at creation time.
Three entry types are maintained:

\begin{itemize}[nosep]
  \item \textbf{Success snippets} $\mathcal{M}_S$: sub-goal action sequences with their precondition snapshots (e.g., ``from $h_t{=}\varnothing$: \texttt{goto fridge} $\to$ \texttt{open fridge} $\to$ \texttt{take mug}'').
  \item \textbf{Failure lessons} $\mathcal{M}_F$: failed actions paired with corrective rules abstracted from the failure context (e.g., ``\texttt{take} returned \texttt{Nothing happens} $\Rightarrow$ violated location precondition'').
  \item \textbf{Verified schemas} $\mathcal{M}_V$: action templates with empirically confirmed precondition--effect pairs and a confidence score (e.g., $\texttt{cool}$: pre~$= \{h_t {\neq} \varnothing,\; \ell_t {=} \texttt{fridge}\}$, eff~$= \{h_t\text{ preserved}\}$, $c = 1.0$).
\end{itemize}

After Round~1 learning, the ALFWorld memory contains 233 entries ($|\mathcal{M}_S|{=}64$, $|\mathcal{M}_F|{=}157$, $|\mathcal{M}_V|{=}12$), frozen before evaluation.
\paragraph{Retrieval.}
Candidate entries are retrieved by partial matching on the goal signature:
\begin{equation}
  \mathcal{M}_{\text{cand}} = \bigl\{\, m \in \mathcal{M} \;\bigm|\; d_\gamma\bigl(\gamma(g),\, \gamma(m)\bigr) \ge \theta \,\bigr\}
  \label{eq:mem_retrieve}
\end{equation}
where $d_\gamma$ is a field-wise score (exact match on task type, fuzzy match on object category and destination) and $\theta$ is a retrieval threshold.

\paragraph{State-Consistent Filtering (SCF).}
Not all goal-relevant entries are appropriate for the current situation: a success snippet recorded with $h_t \neq \varnothing$ should not be injected when the agent's hand is empty.
SCF filters $\mathcal{M}_{\text{cand}}$ against the current state signature:
\begin{equation}
  \mathcal{M}_{\text{filt}} = \{\, m \in \mathcal{M}_{\text{cand}}
  \mid \textsc{Compat}(\sigma(b_t), \sigma(m)) \,\}
  \label{eq:scf}
\end{equation}
An entry $m$ with $\sigma(m) = (l_m, h_m)$ passes iff hand-occupancy matches:
\begin{equation}
  (h_m {=} \varnothing \;\Rightarrow\; h_t {=} \varnothing) \;\wedge\; (h_m {\neq} \varnothing \;\Rightarrow\; h_t {\neq} \varnothing)
  \label{eq:compat_detail}
\end{equation}
This lightweight check eliminates the most common source of mismatched experience injection while preserving recall.

\subsection{Rules-First Arbitration}
\label{sec:arbitration}

After SCF, surviving entries may still conflict with active rules.
The arbitration module resolves conflicts through a two-layer policy:

\begin{enumerate}[nosep]
  \item \textbf{Hard filter}: entries whose suggested actions directly violate a rule are removed (e.g., an entry recommending ``\texttt{put} apple in fridge'' before cooling conflicts with the environment rule that \texttt{cool} is atomic).
  \item \textbf{Soft annotation}: entries with potential but non-contradictory conflicts receive a warning tag, delegating the final judgment to the LLM.
\end{enumerate}

This rules-first priority ensures that grounded action constraints always take precedence over recalled experience, while still allowing the LLM to exercise judgment under ambiguity.

\subsection{Prompt Construction and Decision Procedure}
\label{sec:prompt}

All retrieved knowledge is serialized and concatenated to form the augmented prompt:
\begin{equation}
\begin{aligned}
  p_t = \;& \textsc{Ser}(\text{SysInst}) \oplus \textsc{Ser}(\text{ICL}) \oplus \textsc{Ser}(\mathcal{R}_{\text{active}}) \\
           & \oplus \textsc{Ser}(\gamma_g) \oplus \textsc{Ser}(b_t) \oplus \textsc{Ser}(\mathcal{M}_{\text{arb}}) \\
           & \oplus \textsc{Ser}(H_{1:t-1}) \oplus o_t
\end{aligned}
\label{eq:prompt}
\end{equation}
The baseline prompt is a strict subset omitting $\mathcal{R}_{\text{active}}$, $\gamma_g$, $b_t$, and $\mathcal{M}_{\text{arb}}$ while keeping all shared components identical, so performance differences are primarily attributable to the injected knowledge.
The token overhead of RPMS-injected blocks is approximately 800--1200 tokens per step (see Appendix~\ref{app:prompt}).
The full decision pseudo-code is in Appendix~\ref{app:algo}.

\section{Experimental Setup}
\label{sec:setup}

\subsection{ALFWorld}

We evaluate on ALFWorld v0.3.5~\cite{shridhar2020alfworld}, a text-based embodied benchmark spanning six task types: object retrieval (\texttt{look}, \texttt{pick\_and\_place}), single-step state-change manipulation (\texttt{clean}, \texttt{heat}, \texttt{cool}), and multi-object coordination (\texttt{pick\_two\_obj}), totalling 134 unseen evaluation tasks (out-of-distribution split) and 140 seen tasks used for learning.

\subsection{ScienceWorld}
\label{sec:sw_setup}

To assess cross-environment generalisation, we adapt RPMS to ScienceWorld~\cite{wang2022scienceworld}, a science-experiment simulation with substantially different task structure: longer action horizons (up to 100 steps), a richer action vocabulary (24 commands), and a \emph{continuous} score signal (0--100) rather than binary success.
Following \citet{lin2023swiftsage} and \citet{majumder2024clin}, we evaluate on 26 tasks (excluding 4 electricity tasks) using up to 10 official test variations per task (241 episodes total, as some tasks have fewer than 10 test variations). The primary metric is Average Score (0--100).

\paragraph{Adaptation.}
The three-tier rule hierarchy and SCF mechanism transfer directly; we replace the 6-type goal taxonomy with a \textbf{4-family} classification (F1~Search \& Focus, F2~Measure \& Observe, F3~Transform \& Verify, F4~Long-horizon Procedure) that groups the 26 tasks by their canonical procedure.
The belief state tracks room, inventory occupancy, and a sub-goal progress pointer (details in Appendix~\ref{app:sw_method}).
The SCF compatibility predicate is extended to check instrument availability, room match, and progress alignment.
For ScienceWorld, episodic memory additionally records \textbf{critical failures}---actions that trigger an immediate score of $-100$ (e.g., \texttt{focus on} the wrong object)---with retrieval weight $2.0$ to ensure they are prominently surfaced in future episodes.

\subsection{Model and Decoding}

\paragraph{ALFWorld.} Our primary backbone is Llama~3.1~8B-Instruct~\cite{grattafiori2024llama3}
served via DeepInfra with greedy decoding (temperature $= 0$, max\_tokens $= 256$).
Scaling experiments use Llama~3.1~70B-Instruct and Claude~Sonnet~4.5 with identical hyperparameters.

\paragraph{ScienceWorld.} We use GPT-4~\cite{openai2024gpt4technicalreport} (via the OpenAI API, temperature $= 0$, max\_tokens $= 512$) as the backbone.
The two environments are thus evaluated with different backbone models; cross-environment comparisons should be interpreted as demonstrating architectural transferability rather than model-controlled performance differences.

\paragraph{Evaluation protocol.}
Each task is attempted exactly \emph{once} (single-trial, no retry). Step budgets are $T_{\max} = 50$ (ALFWorld) and $T_{\max} = 100$ (ScienceWorld).
For ALFWorld we report the success rate:
\begin{equation}
  \textit{SR} = \frac{1}{N}\sum_{i=1}^{N}
    \mathbb{1}\bigl[R(s_{T}^{(i)},\, g^{(i)}) = 1\bigr]
  \label{eq:sr}
\end{equation}
For ScienceWorld we report the Average Score (0--100).

\paragraph{Learning phase.}
During the learning phase the agent runs the seen-split tasks under the Rules-only configuration and accumulates episodic memory $\mathcal{M}$ with deduplication. After Round~1, the ALFWorld memory contains 233 entries ($|\mathcal{M}_S|{=}64$, $|\mathcal{M}_F|{=}157$, $|\mathcal{M}_V|{=}12$). This memory is then \emph{frozen} before evaluation and shared across all ablation conditions; no online learning occurs during evaluation.

\subsection{Baseline}

Our controlled baseline is \textbf{ReAct}~\cite{yao2023react} with MPO-aligned~\cite{xiong2025mpo} system prompts and ICL examples. This baseline shares \emph{all} prompt components with RPMS---system instruction, ICL example, action format, and decoding parameters---except the knowledge injection blocks ($\mathcal{R}_{\text{active}}$, $\gamma_g$, $b_t$, $\mathcal{M}_{\text{arb}}$). Observed performance differences in this controlled comparison should therefore be interpreted mainly through the presence or absence of the injected knowledge blocks (Sec.~\ref{sec:prompt}).

Published numbers from prior papers are used only as qualitative context in Appendix~\ref{app:published_context}, because their backbones, prompts, trial budgets, or training procedures differ from ours.

\subsection{Ablation Configurations}

To disentangle the contributions of rules and memory, we design a $2{\times}2$ factorial ablation with four conditions: (A) Baseline = no rules, no memory; (B) Rules-only = rules on, memory off; (C) Memory-only = rules off, memory on; and (D) Full RPMS = both on. All four conditions use the same LLM backbone, decoding parameters, system instruction, ICL example, and---where applicable---the same frozen 233-entry memory; they differ \emph{only} in which knowledge blocks are present in the prompt. This factorial design enables us to measure (i)~the marginal effect of rules (A$\to$B), (ii)~the marginal effect of memory (A$\to$C), and (iii)~any interaction effect in the full system (D vs.\ expected additive A+B+C).

\section{Results}
\label{sec:results}

\subsection{Main Results (ALFWorld)}

\begin{table}[!t]
\centering
\caption{Controlled same-protocol results on 134 unseen ALFWorld tasks (single-trial). $\Delta$: absolute improvement over each model's ReAct baseline.}
\label{tab:main_results}
\small
\begin{tabular}{llcc}
\toprule
\textbf{Model} & \textbf{Method} & \textbf{Succ.\ (\%)} & $\boldsymbol{\Delta}$ \\
\midrule
\multirow{2}{*}{\makecell[l]{Llama 3.1\\8B}}
  & ReAct (Baseline)          & 35.8 & --- \\
  & \textbf{RPMS (ours)}      & \textbf{59.7} & \textbf{+23.9} \\
\midrule
\multirow{2}{*}{\makecell[l]{Llama 3.1\\70B}}
  & ReAct              & 72.4 & --- \\
  & \textbf{RPMS}      & \textbf{88.1} & \textbf{+15.7} \\
\midrule
\multirow{2}{*}{\makecell[l]{Claude\\Sonnet 4.5}}
  & ReAct              & 86.6 & --- \\
  & \textbf{RPMS}      & \textbf{98.5} & \textbf{+11.9} \\
\bottomrule
\end{tabular}
\end{table}

Table~\ref{tab:main_results} reports the controlled comparison under matched prompts, decoding settings, and evaluation budget.
With Llama~3.1~8B, RPMS achieves \textbf{59.7\%} success (80/134), a +23.9~pp improvement over ReAct (48/134).
Rule retrieval alone accounts for +14.9~pp of this gain and is statistically significant ($p = 0.008$, Appendix~\ref{app:significance}); the additional contribution from state-consistent memory reflects the conditional-usefulness finding described in Sec.~\ref{sec:discussion}.
The advantage persists at larger scales: +15.7~pp with 70B and +11.9~pp with Claude~Sonnet~4.5, the latter reaching 98.5\%.

\subsection{Per-Task-Type Analysis}
\label{sec:per_task}

\begin{table*}[!t]
\centering
\caption{Per-task-type success rate (\%) on 134 unseen ALFWorld tasks (Llama~3.1~8B). N = number of tasks per type. Best in \textbf{bold}.}
\label{tab:per_task}
\small
\begin{tabular}{lcccccc|c}
\toprule
\textbf{Method} & \textbf{Look} (18) & \textbf{Place} (24) & \textbf{Clean} (31) & \textbf{Cool} (21) & \textbf{Heat} (23) & \textbf{Two} (17) & \textbf{All} (134) \\
\midrule
Baseline         & 55.6 & 33.3 & 29.0 & 42.9 & 30.4 & 29.4 & 35.8 \\
Memory-only      & 11.1 & 70.8 & 38.7 & 52.4 & 26.1 & 41.2 & 41.0 \\
Rules-only       & 33.3 & 75.0 & 41.9 & 61.9 & 52.2 & 35.3 & 50.7 \\
\textbf{Full RPMS} & \textbf{66.7} & \textbf{79.2} & \textbf{54.8} & 47.6 & \textbf{65.2} & \textbf{41.2} & \textbf{59.7} \\
\bottomrule
\end{tabular}
\end{table*}

Table~\ref{tab:per_task} reveals several patterns.
First, RPMS achieves the highest success across five of six task types.
The exception is \textbf{Cool}, where Rules-only (61.9\%) outperforms Full~RPMS (47.6\%); we attribute this to a conflict between stored failure lessons and environment-specific action constraints (see Limitations).

Second, the largest absolute gains appear on \textbf{Heat} (+34.8~pp over Baseline) and \textbf{Place} (+45.9~pp), both requiring strict multi-step procedures that benefit from explicit rule injection.

Third, Memory-only dramatically improves Place (33.3\%$\to$70.8\%) but \emph{degrades} Look (55.6\%$\to$11.1\%), indicating that unfiltered experiential knowledge can harm performance on simple tasks.
Full~RPMS recovers Look to 66.7\%, confirming that rules and arbitration mitigate memory-induced degradation.

\subsection{Ablation Study: Rules vs.\ Memory}
\label{sec:ablation}

\begin{table}[!t]
\centering
\caption{2$\times$2 ablation (Llama~3.1~8B, 134 unseen ALFWorld tasks).}
\label{tab:ablation}
\small
\begin{tabular}{lcccc}
\toprule
\textbf{Condition} & \textbf{Rules} & \textbf{Mem.} & \textbf{Succ.} & \textbf{Rate} \\
\midrule
(A) Baseline    & OFF & OFF & 48/134  & 35.8\% \\
(C) Memory-only & OFF & ON  & 55/134  & 41.0\% \\
(B) Rules-only  & ON  & OFF & 68/134  & 50.7\% \\
(D) Full RPMS   & ON  & ON  & 80/134  & \textbf{59.7\%} \\
\bottomrule
\end{tabular}
\end{table}

Table~\ref{tab:ablation} and Fig.~\ref{fig:ablation_combined} (left) report the full 2$\times$2 ablation.
Rules independently contribute +14.9~pp (A$\to$B), Memory contributes +5.2~pp (A$\to$C), while the combined system achieves +23.9~pp (A$\to$D).
The combined gain (+23.9~pp) exceeds the sum of the two single-factor contributions (+20.1~pp), suggesting a positive interaction between rule injection and memory on this benchmark. Arbitration analysis and efficiency results are in Appendix~\ref{app:extended_analysis}.

\subsection{Arbitration Analysis}
\label{sec:arbitration_analysis}

\begin{table}[!t]
\centering
\caption{Arbitration mode comparison (Rules ON + Memory ON, 8B).}
\label{tab:arbitration}
\small
\begin{tabular}{lcc}
\toprule
\textbf{Arb.\ Mode} & \textbf{Succ.} & \textbf{Rate (\%)} \\
\midrule
None               & 64/134 & 47.8 \\
Soft-only          & 71/134 & 53.0 \\
Hard-only          & 74/134 & 55.2 \\
\textbf{Hard+Soft} & \textbf{80/134} & \textbf{59.7} \\
\bottomrule
\end{tabular}
\end{table}

Table~\ref{tab:arbitration} shows that naively combining rules and memory without conflict resolution is weaker than Rules-only (47.8\% vs.\ 50.7\%).
Hard filtering removes direct contradictions; adding soft annotation yields a further gain to 59.7\%, consistent with the claim that arbitration matters when both knowledge sources are active.

\subsection{Generalization to ScienceWorld}
\label{sec:sw_results}

\begin{table}[!t]
\centering
\caption{2$\times$2 ablation on ScienceWorld (GPT-4, 241 episodes). Metric: Average Score (0--100).}
\label{tab:sw_ablation}
\small
\begin{tabular}{lccc}
\toprule
\textbf{Condition} & \textbf{Rules} & \textbf{Mem.} & \textbf{Avg.\ Score} \\
\midrule
(A) Baseline    & OFF & OFF & 44.9 \\
(C) Memory-only & OFF & ON  & 46.0 \\
(B) Rules-only  & ON  & OFF & 51.3 \\
(D) Full RPMS   & ON  & ON  & \textbf{54.0} \\
\bottomrule
\end{tabular}
\end{table}

\begin{figure}[!t]
  \centering
  \includegraphics[width=\columnwidth]{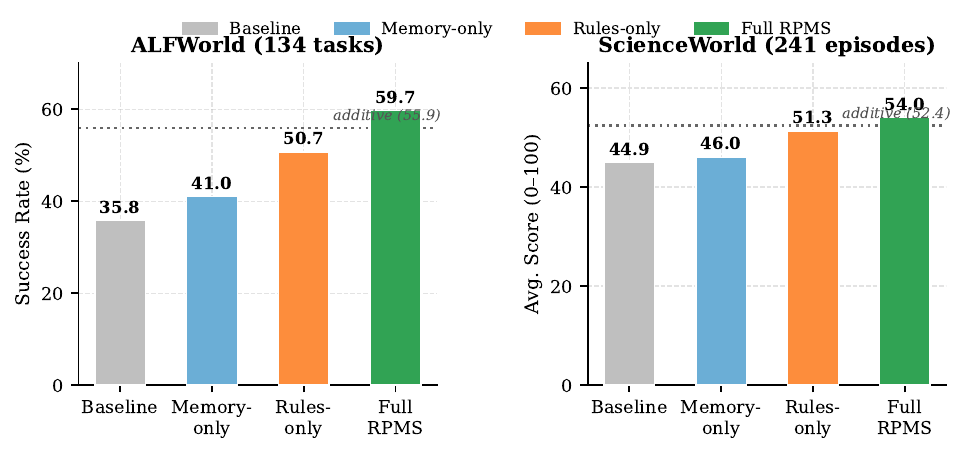}
  \caption{2$\times$2 ablation results on ALFWorld (left, success rate \%; backbone: Llama~3.1~8B) and ScienceWorld (right, avg.\ score 0--100; backbone: GPT-4). Dotted lines mark the additive-expectation baseline in each environment.}
  \label{fig:ablation_combined}
\end{figure}

Table~\ref{tab:sw_ablation} and Fig.~\ref{fig:ablation_combined} (right) report the ScienceWorld 2$\times$2 ablation.
Full RPMS achieves an average score of \textbf{54.0} against a ReAct baseline of 44.9 (+9.1 points), with rules contributing the larger share (+6.4 points, A$\to$B) and memory adding a further +1.1 (A$\to$C).
The combined system outperforms both single-factor conditions, consistent with the same rule--memory interaction observed on ALFWorld.

Comparing across environments, two consistent patterns emerge. First, \textbf{rules remain the dominant contributor in both settings}, consistent with the hypothesis that performance is constrained by missing action semantics rather than by reasoning ability alone. Second, the \textbf{relative memory contribution is smaller in ScienceWorld} (+1.1 vs.\ +5.2~pp on ALFWorld), which we attribute to the noisier continuous score signal and longer horizons making it harder to identify exactly which sub-sequence of an episode should be stored or recalled. The critical-failure mechanism partially compensates by prioritising the most consequential negative experiences (Sec.~\ref{sec:sw_setup}).
Note that ScienceWorld experiments use GPT-4 as the backbone while ALFWorld experiments use Llama~3.1~8B; the consistent direction of gains across both conditions is therefore evidence of architectural transferability rather than a controlled model comparison.
The impact of SCF also differs between environments: disabling it costs --6.0~pp on ALFWorld Full~RPMS but less on ScienceWorld, where the progress-pointer compatibility check already filters a substantial fraction of mismatched entries---suggesting that richer state signatures can partially substitute for the hand-occupancy check used in ALFWorld.

Published results from prior systems on both benchmarks are provided for qualitative context in Appendix~\ref{app:published_context}.

\section{Discussion}
\label{sec:discussion}

Two observations are most relevant to our central thesis---that memory is reliably helpful only when filtered by current state and grounded by explicit action constraints.
First, rules account for the larger single-factor gain (+14.9~pp on ALFWorld, +6.4 on ScienceWorld), consistent with the view that performance is constrained by missing action semantics at least as much as by reasoning ability~\cite{ahn2022saycan,singh2023progprompt}.
Second, and crucially, memory is \emph{not} universally beneficial: Memory-only improves Place (33.3\%$\to$70.8\%) yet degrades Look (55.6\%$\to$11.1\%)---consistent with the claim that unfiltered retrieved knowledge introduces inapplicable advice~\cite{zhang2023rememberer}.
Full~RPMS recovers Look to 66.7\% via arbitration, providing direct evidence that the conditional-usefulness claim holds: the same memory that is harmful without state filtering and rule grounding becomes a net positive with it.

The cross-environment pattern reinforces this thesis. In ScienceWorld, rules again dominate the single-factor gains and memory contributes only conditionally; the critical-failure mechanism partially offsets the noisier score signal by prioritising the most consequential negative experiences over ordinary failure lessons. The consistent ordering across two environments with different action spaces, score signals, and backbone models strengthens the claim that state-aware knowledge mediation is the operative mechanism rather than any environment-specific feature of RPMS.
\section{Conclusion}
We presented RPMS, grounded in the observation that in closed-world embodied planning, \emph{external knowledge is only reliably useful when it is both executable and state-consistent at decision time}. P1 (invalid action generation) and P2 (state drift) represent the two ways this condition can fail; RPMS addresses both through a conflict-managed combination of rule retrieval, lightweight belief tracking, and state-filtered episodic memory. On 134 unseen ALFWorld tasks, RPMS achieves 59.7\% single-trial success with Llama~3.1~8B and 98.5\% with Claude Sonnet~4.5. The +23.9~pp combined gain substantially exceeds the sum of individual contributions---but the more important finding is that episodic memory alone degrades some task types and only becomes a stable net positive when grounded by explicit action constraints. Across two structurally distinct environments the evidence consistently supports this architectural thesis: rules dominate single-factor gains, memory is conditional rather than universally helpful, and arbitration matters when both sources are active. Extending state-aware knowledge mediation to domains where rule authoring can be automated is the principal direction for future work.

\section*{Limitations}
We identify four directions where RPMS can be extended.
First, the episodic memory module requires prior interaction experience to be useful; 
it cannot generate applicable guidance from scratch. 
Although memory gains saturate quickly in practice and the learning phase is 
lightweight, developing mechanisms to bootstrap useful memory entries without 
any prior trajectories remains an open challenge.
Second, state-consistent filtering uses hand-occupancy compatibility as a 
lightweight proxy for full precondition equivalence---a deliberate trade-off 
favouring recall over precision that leaves room for finer-grained filtering.
Third, the rule manual's environment-specific tier (Tier 3) is domain-dependent; 
extending RPMS to new domains requires authoring or automatically inducing 
environment-specific action semantics from environment feedback~\cite{gao2024ragsurvey}, 
while the universal and domain-procedure tiers transfer largely unchanged, 
as demonstrated by the ScienceWorld adaptation.
Fourth, the current arbitration module operates at the entry level; finer-grained 
conflict resolution---distinguishing an entry's action-sequence advice from 
embedded location cues---could recover additional gains in cases where the two 
are mixed.

\section*{Ethical Considerations}
RPMS is designed to improve reliability in simulated embodied environments 
and is not intended for deployment in safety-critical real-world settings 
without additional safeguards. We identify two sources of potential risk 
specific to this architecture. First, the rule manual is human-authored: 
incomplete or incorrect rules will cause the agent to systematically 
generate invalid actions in ways that may be difficult to diagnose, since 
the rules-first arbitration policy propagates rule errors throughout the 
decision pipeline. Second, episodic memory is accumulated from prior 
interaction trajectories: if the seen-task distribution is biased or 
contains erroneous episodes, the memory module will surface and reinforce 
those biases during retrieval, particularly in out-of-distribution settings 
where filtering may be less effective. Beyond these architecture-specific 
risks, the rules-first design reduces the agent's reliance on open-ended 
LLM judgment, which is beneficial in well-specified environments but may 
produce unpredictable behavior in novel situations not covered by the 
current rule tiers. We therefore recommend that any real-world extension 
of RPMS be accompanied by systematic rule auditing, trajectory-level 
memory review, human oversight, and sandboxed validation before deployment.

\bibliography{custom}

\appendix

\section{Supplementary Method Details}
\label{app:method_appendix}

\subsection{Pseudo-code}
\label{app:algo}

\begin{algorithm}[h]
\caption{RPMS Single-Step Decision}
\label{alg:rpms}
\footnotesize
\begin{algorithmic}[1]
\REQUIRE Observation $o_t$, goal $g$, prior state $b_{t-1}$, history $H_{1:t-1}$
\ENSURE Action $a_t$, updated belief $b_t$
\STATE $b_t \leftarrow f_{\textsc{upd}}(b_{t-1}, a_{t-1}, o_t)$
\STATE $\sigma_t \leftarrow (\ell_t, h_t)$; $\gamma_g \leftarrow (g_{\textit{type}}, g_{\textit{obj}}, g_{\textit{dest}})$
\STATE $\mathcal{R}_{\text{active}} \leftarrow \mathcal{R}_U \cup \mathcal{R}_D(g_{\textit{type}}) \cup \mathcal{R}_E$
\STATE $\mathcal{M}_{\text{cand}} \leftarrow \{m \in \mathcal{M} \mid d_\gamma(\gamma_g, \gamma(m)) \ge \theta\}$
\STATE $\mathcal{M}_{\text{filt}} \leftarrow \{m \in \mathcal{M}_{\text{cand}} \mid \textsc{Compat}(\sigma_t, \sigma(m))\}$
\STATE $\mathcal{M}_{\text{arb}} \leftarrow \{\phi(m) \mid m \in \mathcal{M}_{\text{filt}},\; \phi(m) \neq \bot\}$
\STATE $p_t \leftarrow \textsc{Build}(\text{SysInst}, \text{ICL}, \mathcal{R}_{\text{active}}, \gamma_g,$
\STATE \hspace{1.35em}$b_t, \mathcal{M}_{\text{arb}}, H_{1:t-1}, o_t)$
\STATE $a_t \leftarrow \textsc{LLM}(p_t)$
\RETURN $a_t, b_t$
\end{algorithmic}
\end{algorithm}

\subsection{ScienceWorld Adaptation Details}
\label{app:sw_method}

\paragraph{4-Family GoalSpec.}
The 26 evaluated ScienceWorld tasks are partitioned into four families: F1~Search \& Focus (4 tasks), F2~Measure \& Observe (3), F3~Transform \& Verify (7), and F4~Long-horizon Procedure (12). Domain rules are filtered by family, analogously to ALFWorld's per-type domain rules.

\paragraph{BeliefState.}
The ScienceWorld belief state tracks room, inventory set, visible objects, state flags (e.g., \texttt{water} $\to$ \texttt{gas}), and a sub-goal progress pointer. All updates are deterministic regex-based parses of observation text---no LLM call is involved.

\paragraph{SCF compatibility.}
$\textsc{Compat}_{\textit{SW}}$ verifies three conditions:
(1)~\emph{instrument compatibility} (hard)---if the stored snippet required an instrument, the agent must currently hold one;
(2)~\emph{room compatibility} (hard)---if the entry specifies a room, the agent must be in that room;
(3)~\emph{progress compatibility} (bounded lag)---the entry's minimum progress pointer must not exceed the current pointer by more than one step.

\paragraph{Critical failures.}
Actions that trigger an immediate score of $-100$ (e.g., \texttt{focus on} the wrong object) are stored as \textsc{Critical\_Failure} entries with retrieval weight $2.0$ (vs.\ $1.0$ for ordinary failures) and marked \texttt{[CRITICAL]} in the prompt.

\section{Published Results for Context}
\label{app:published_context}

Tables~\ref{tab:published_alfworld} and~\ref{tab:published_sw} list published results on ALFWorld and ScienceWorld for qualitative reference. Direct comparison with our results is \emph{not} intended: methods differ in backbone, training regime, trial budget, and evaluated subset.

\begin{table*}[!t]
\centering
\caption{Published ALFWorld results for qualitative context only. Methods differ in backbone, training regime, and trial budget.}
\label{tab:published_alfworld}
\scriptsize
\setlength{\tabcolsep}{3pt}
\begin{tabular}{@{}lp{3.4cm}p{4.0cm}p{1.8cm}lr@{}}
\toprule
\textbf{Method} & \textbf{Backbone / Agent} & \textbf{Regime} & \textbf{Trials} & \textbf{Reference} & \textbf{SR} \\
\midrule
BUTLER8    & Trained policy (IL / seq2seq) & Trained                            & best-of-8  & \citet{shridhar2020alfworld} & 37 \\
ReAct      & PaLM-540B / text-davinci-002  & Prompt-only                        & single-run & \citet{yao2023react}         & 70.9/78.4 \\
Reflexion  & GPT-3.5                       & Prompt-only (iterative reflection) & 11 iter.   & \citet{shinn2023reflexion}   & 97 \\
RAFA       & GPT-4                         & Prompt-only (iterative replanning) & 8 iter.    & \citet{liu2023rafa}          & 99 \\
A3T        & Trained (self-training)       & Trained                            & single-run & \citet{yang2024actre}        & 86--96 \\
\midrule
\textbf{Ours} & \textbf{Llama-3.1-8B}      & \textbf{Prompt-only (rule+mem)}    & \textbf{single-run} & --- & \textbf{59.7} \\
\textbf{Ours} & \textbf{Llama-3.1-70B}     & \textbf{Prompt-only (rule+mem)}    & \textbf{single-run} & --- & \textbf{88.1} \\
\textbf{Ours} & \textbf{Claude Sonnet 4.5} & \textbf{Prompt-only (rule+mem)}    & \textbf{single-run} & --- & \textbf{98.5} \\
\bottomrule
\end{tabular}
\end{table*}

\begin{table*}[!t]
\centering
\caption{Published ScienceWorld results for qualitative context only. Methods differ in backbone, training regime, protocol, and evaluated subset.}
\label{tab:published_sw}
\scriptsize
\setlength{\tabcolsep}{3pt}
\begin{tabular}{@{}lp{3.2cm}p{4.0cm}p{1.8cm}lr@{}}
\toprule
\textbf{Method} & \textbf{Backbone} & \textbf{Regime} & \textbf{Split} & \textbf{Reference} & \textbf{Score} \\
\midrule
SwiftSage: ReAct     & GPT-4                & Prompt-only (few-shot)             & 30 tasks & \citet{lin2023swiftsage}    & 36.4 \\
SwiftSage: Reflexion & GPT-4                & Prompt-only (iterative reflection) & 30 tasks & \citet{lin2023swiftsage}    & 45.3 \\
Retrospex: IL-T5     & Flan-T5-large        & Trained (IL)                       & 30 tasks & \citet{xiang-etal-2024-retrospex} & 48.8 \\
Retrospex            & Flan-T5-large + RL   & Trained (IL+RL)                    & 30 tasks & \citet{xiang-etal-2024-retrospex} & 56.0 \\
CLIN                 & GPT-4                & Continual memory update            & 30 tasks & \citet{majumder2024clin}    & 59.8 \\
SwiftSage            & SWIFT + SAGE (GPT-4) & Hybrid (BC + prompting)            & 30 tasks & \citet{lin2023swiftsage}    & 84.7 \\
\midrule
\textbf{Ours (RPMS)} & \textbf{GPT-4}      & \textbf{Prompt-only (rule+mem)}    & \textbf{26 tasks, up to 10 var.\ (241 ep.)} & --- & \textbf{54.0} \\
\bottomrule
\end{tabular}
\end{table*}

\section{Extended Analysis}
\label{app:extended_analysis}

\subsection{Execution Efficiency}

\begin{table}[!t]
\centering
\caption{Average execution steps (Llama~3.1~8B, 134 ALFWorld tasks).}
\label{tab:efficiency}
\small
\begin{tabular}{lcc}
\toprule
\textbf{Condition} & \textbf{Avg Steps (All)} & \textbf{Avg Steps (Succ.)} \\
\midrule
Baseline      & 37.3 & 14.6 \\
Memory-only   & 35.1 & 13.8 \\
Rules-only    & 32.1 & 14.7 \\
Full RPMS     & \textbf{30.9} & \textbf{14.0} \\
\bottomrule
\end{tabular}
\end{table}

RPMS reduces overall average steps from 37.3 to 30.9 (17.2\% reduction, Table~\ref{tab:efficiency}).
Among successful tasks all methods require $\sim$14 steps, indicating comparable inherent task difficulty.
The reduction is driven by fewer wasted steps on failed tasks: RPMS fails faster via early error detection rather than exhausting the 50-step budget.

\subsection{Failure Mode Analysis}

\begin{table}[!t]
\centering
\caption{Failure mode distribution among failed tasks (ALFWorld).}
\label{tab:failure_modes}
\small
\begin{tabular}{lcccc}
\toprule
\textbf{Failure Mode} & \textbf{Base} & \textbf{Mem} & \textbf{Rules} & \textbf{RPMS} \\
\midrule
Timeout (50 steps) & 86 & 79 & 27 & 18 \\
Env-done (wrong)   &  0 &  0 & 39 & 36 \\
\midrule
Total failed & 86 & 79 & 66 & 54 \\
\bottomrule
\end{tabular}
\end{table}

Table~\ref{tab:failure_modes} reveals a qualitative shift in failure behavior.
Without rules, \emph{all} failures are timeouts---the agent wanders for 50 steps.
With rules, a majority of failures become early terminations from incorrect actions.
These counts are consistent with rules making the agent more \emph{decisive}: it commits more quickly, sometimes incorrectly, but avoids the timeout trap and achieves a much higher success rate overall.

\subsection{Learning Curve}
\label{sec:learning_curve}

We investigate how accumulated experience affects performance.
The agent learns on 140 seen tasks, accumulating episodic memory through smart deduplication merging across 3 rounds (snapshots of 233, 331, and 449 entries), then evaluates on the 134 unseen tasks with frozen memory.

\begin{table}[!t]
\centering
\caption{Learning curve (Llama~3.1~8B, 134 unseen ALFWorld tasks).}
\label{tab:learning_curve}
\small
\resizebox{\columnwidth}{!}{%
\begin{tabular}{lccc}
\toprule
\textbf{Round} & \textbf{Entries} & \textbf{Mem-only (\%)} & \textbf{Rules+Mem (\%)} \\
\midrule
R0 (no mem) & 0   & 35.8 & 50.7 \\
R1          & 223 & 38.8 & 53.0 \\
R2          & 331 & 41.8 & 54.5 \\
R3          & 449 & 42.5 & 56.0 \\
\bottomrule
\end{tabular}
}%
\end{table}

\begin{figure}[!t]
  \centering
  \includegraphics[width=\columnwidth]{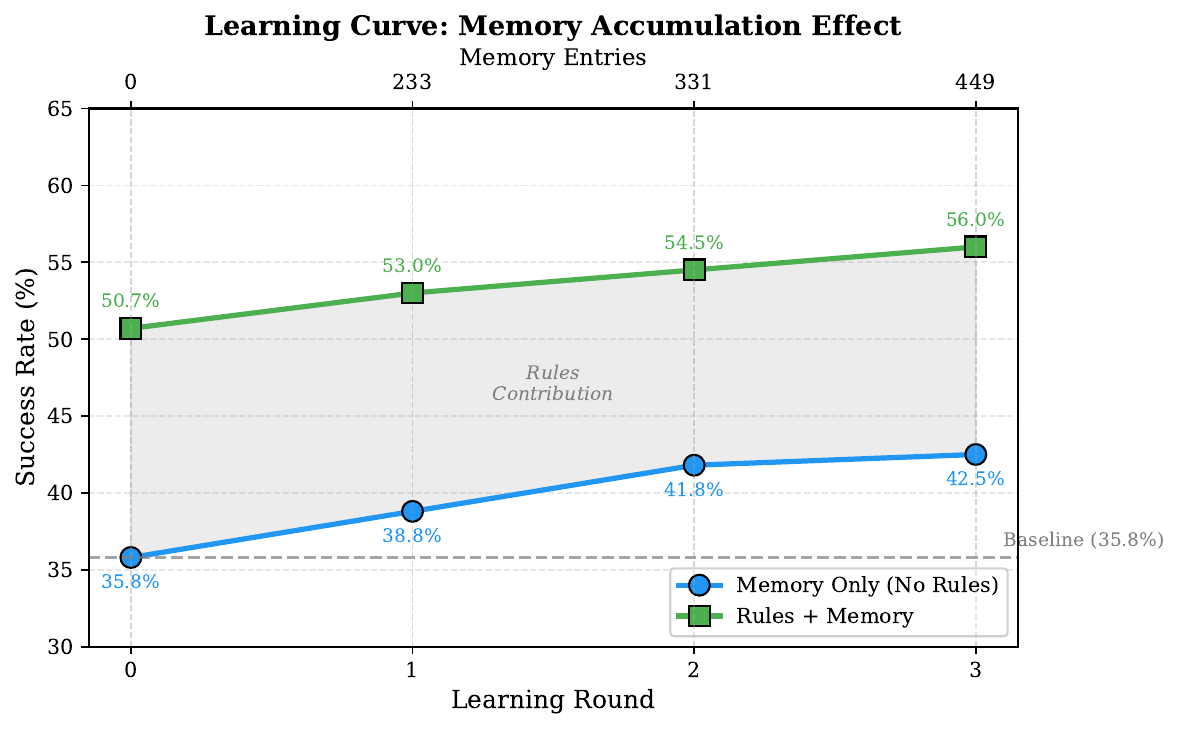}
  \caption{Learning curve: success rate vs.\ learning rounds for Memory-only and Rules+Memory configurations.}
  \label{fig:learning_curve}
\end{figure}

Table~\ref{tab:learning_curve} and Fig.~\ref{fig:learning_curve} show two findings.
First, \textbf{memory provides consistent but saturating gains}: Memory-only improves from 35.8\% to 42.5\%, with most gains in the first two rounds.
Second, \textbf{rules amplify memory effectiveness}: the gap between Rules+Memory and Memory-only is stable at $\sim$13--14~pp across all rounds, confirming orthogonal benefits.

Disabling state-consistent filtering costs --3.7~pp on Memory-only and --6.0~pp on Full~RPMS, confirming that state-inconsistent memories are more damaging when combined with rules.

\FloatBarrier
\section{Complete Prompt Examples}
\label{app:prompt}

We show the complete prompts for a representative \texttt{pick\_heat\_then\_place} ALFWorld task (``put a hot apple in fridge'') under both conditions at step~1, and the RPMS injection for a ScienceWorld F3 task (``boil water'').

\subsection{ALFWorld: Baseline Prompt}

The baseline prompt consists of three blocks: system instruction, ICL example, and current observation. The ICL example (``heat egg $\to$ diningtable'') is a fixed 1-shot demonstration drawn from a \emph{different} task; it is identical across all conditions.

\begin{lstlisting}[caption={System Instruction (shared by Baseline and RPMS)}]
Interact with a household to solve a task. Imagine
you are an intelligent agent in a household
environment and your target is to perform actions
to complete the task goal. [...]
The available actions are:
1. go to {recep}
2. take {obj} from {recep}
3. put {obj} in/on {recep}
4. open {recep}     5. close {recep}
6. toggle {obj} {recep}
7. clean {obj} with {recep}
8. heat {obj} with {recep}
9. cool {obj} with {recep}
Reminder:
1. The action must be chosen from the given
   available actions.
2. Think when necessary, try to act directly more.
\end{lstlisting}

\begin{lstlisting}[caption={ICL Example (1-shot, heat task)}]
You are in the middle of a room. [...]
Your task is to: heat some egg and put it in
diningtable.
Thought: To solve the task, I need to find an egg,
heat it with microwave, then put it in diningtable.
First I need to find an egg. [...]
Action: open fridge 1
Obs: You open the fridge 1. [...]

Action: go to countertop 3
Obs: On the countertop 3, you see a egg 2 [...]

Action: take egg 2 from countertop 3
Obs: You pick up the egg 2.

Action: go to microwave 1
Obs: The microwave 1 is closed.

Action: heat egg 2 with microwave 1
Obs: You heat the egg 2 using the microwave 1.

Action: go to diningtable 1
Obs: On the diningtable 1, you see [...]

Action: put egg 2 in/on diningtable 1
Obs: You put the egg 2 in/on the diningtable 1.
\end{lstlisting}

\begin{lstlisting}[caption={Observation (step 1, identical across conditions)}]
You are in the middle of a room. Looking quickly
around you, you see a cabinet 6, a cabinet 5,
a cabinet 4, a cabinet 3, a cabinet 2, a cabinet 1,
a coffeemachine 1, a countertop 3, a countertop 2,
a countertop 1, a drawer 3, a drawer 2, a drawer 1,
a fridge 1, a garbagecan 1, a microwave 1,
a shelf 3, a shelf 2, a shelf 1, a sinkbasin 1,
a stoveburner 4, a stoveburner 3, a stoveburner 2,
a stoveburner 1, and a toaster 1.
Your task is to: put a hot apple in fridge.
\end{lstlisting}

\subsection{ALFWorld: RPMS Additional Blocks}

RPMS appends the following structured knowledge to the system prompt. The system instruction, ICL example, and observation remain identical.

\begin{lstlisting}[caption={RPMS: State + Goal + Rules injection ($\sim$3100 chars)}]
[Current State]
  Holding: nothing

[Task Goal]
  Goal: put a hot apple in fridge.
  Type: pick_heat_then_place_in_recep
  Current sub-goal: find_object - apple
  Progress: 0%

[RULES - CRITICAL, Always Follow]
=== Universal Principles ===
  [U-SEARCH-001] When target object is not visible,
    systematically search starting from most likely
    locations
  [U-SEARCH-004] Check open spaces (countertop,
    table) before closed containers
  [U-STATE-001] Before any action, verify
    preconditions against current BeliefState
  [U-FAIL-002] If 'Nothing happens', the action's
    precondition was not met

=== Domain Knowledge ===
  [D-LOC-001] Food items commonly in: fridge,
    countertop, diningtable, cabinet
  [D-TRANS-001] State-change actions require being
    at tool location with object in hand
  [D-TRANS-003] Sequence: find -> take -> go to
    tool -> transform -> go to dest -> place

=== Environment-Specific ===
  HEAT:
    Syntax: heat <object> with microwave 1
    CRITICAL: Single-step atomic action. Object
    stays in hand. Do NOT put object in microwave.

[Task-Specific Rules]
  - S-002: Hands empty - can pick up an object
  - D-HEAT-001: 'heat X with microwave' keeps X
    in hand - do NOT put X in microwave to heat it
\end{lstlisting}

Note: for this particular task, episodic memory was empty (no matching entries found for goal signature \texttt{pick\_heat:apple:hot:fridge}).

\subsection{ALFWorld: RPMS Trajectory (10 steps, SUCCESS)}

\begin{lstlisting}[caption={Full RPMS trajectory for ``put a hot apple in fridge''}]
Step 1: [Thought] I need to find an apple. Food
  items are commonly found in fridge, countertop...
  Action: open fridge 1
  Obs: Nothing happens.

Step 2: [Thought] Fridge must already be open.
  Action: take apple 1 from fridge 1
  Obs: Nothing happens.

Step 3: [Thought] Apple not in fridge. Check
  countertops per D-LOC-001.
  Action: go to countertop 1
  Obs: On countertop 1: apple 2, apple 1, mug 3,
       potato 3, spatula 1.

Step 4: Action: take apple 1 from countertop 1
  Obs: You pick up the apple 1.

Step 5: [Thought] Now holding apple 1. Per
  D-TRANS-003: go to microwave, then heat.
  Action: go to microwave 1
  Obs: The microwave 1 is closed.

Step 6: Action: open microwave 1
  Obs: You open microwave 1. In it: mug 1.

Step 7: [Thought] Per D-HEAT-001: heat is atomic,
  object stays in hand.
  Action: heat apple 1 with microwave 1
  Obs: You heat the apple 1 using microwave 1.

Step 8: Action: go to fridge 1
  Obs: The fridge 1 is closed.

Step 9: Action: open fridge 1
  Obs: You open fridge 1. In it: bowl 1, cup 2,
       cup 1, egg 1, lettuce 1, ...

Step 10: Action: put apple 1 in/on fridge 1
  Obs: You put the apple 1 in/on the fridge 1.

Result: SUCCESS in 10 steps.
\end{lstlisting}

\subsection{ScienceWorld: RPMS Injection (boil task, step 1)}

\begin{lstlisting}[caption={RPMS: State + Goal + Rules injection (ScienceWorld boil task, step 1, $\sim$2900 chars)}]
[Current State]
  Room: unknown  Carrying: nothing  Progress: step 0

[Current Objective] find water

[RULES - Tier 1: Universal Principles]
  U-SEARCH-001: When looking for an object, use
    'look around' before moving elsewhere
  U-INSTR-001: Find instrument BEFORE target if
    required by task
  U-FAIL-001: If 'No known action', rephrase using
    exact object names from the observation
  U-VERIFY-001: After state-change, confirm result

[RULES - Tier 2: Domain (F3_transform_verify)]
  Canonical procedure:
    1. find target substance
    2. focus on TARGET SUBSTANCE (NOT device)
    3. pick up or prepare target
    4. go to apparatus location
    5. execute transform action
    6. observe result (verify state change)
  F3-FOCUS-001: For melt/freeze/boil, 'focus on'
    TARGET SUBSTANCE (e.g., water), NEVER on
    devices (stove) or containers (metal pot)
  WARNING: focus on wrong object -> score=-100

[RULES - Tier 3: Environment (ScienceWorld)]
  E-ACT-005: 'focus on <object>' USE WITH CAUTION
  E-SEM-003: wrong object -> score=-100, terminated

[Past Experience - Learned from History]
  OK: F3: focus on substance in metal pot ->
    move metal pot to stove -> activate stove
  AVOID [CRITICAL]: focus on stove
    Rule: verify object identity matches task
      target before executing focus
\end{lstlisting}

\section{Hierarchical Rule Manual}
\label{app:rules}

The complete rule manual is a YAML file with three tiers. We reproduce representative rules from each tier.

\subsection{Universal Rules (Tier 1)}

These rules apply to all tasks regardless of domain.

\begin{lstlisting}[caption={Selected universal rules}]
search_strategy:
  U-SEARCH-001: When target object is not visible,
    systematically search starting from most likely
    locations.
  U-SEARCH-004: Check open spaces (countertop,
    table) before closed containers.
  U-SEARCH-005: When searching location type X
    (e.g., desk), check ALL instances (desk 1,
    desk 2, ...) before moving to next type.

failure_recovery:
  U-FAIL-001: If the same action fails twice
    consecutively, try an alternative action.
  U-FAIL-002: If 'Nothing happens', the action's
    precondition was not met - diagnose which
    precondition failed.

state_management:
  U-STATE-001: Before any action, verify
    preconditions against current BeliefState.
  U-STATE-003: If holding an object, complete
    current placement before searching for another.
\end{lstlisting}

\subsection{Domain Rules (Tier 2)}

These rules encode environment-specific procedural knowledge.

\begin{lstlisting}[caption={Selected domain rules (ALFWorld household)}]
container_interaction:
  D-CONT-001: Closed containers must be opened
    before accessing contents.
  D-CONT-002: Only one object can be held at a time.

object_locality:
  D-LOC-001 (food): fridge -> countertop ->
    diningtable -> cabinet
  D-LOC-003 (electronics): desk -> shelf ->
    drawer -> sidetable -> bed -> sofa

transformation_pattern:
  D-TRANS-001: State-change actions require being
    at tool location with object in hand.
  D-TRANS-003: Sequence: find -> take -> go to
    tool -> transform -> go to dest -> place.

examination_pattern:
  D-EXAM-001: To examine under light, FIRST take
    the target object, THEN go to the desklamp.
  D-EXAM-003: You examine the OBJECT, not the
    light source.
\end{lstlisting}

\subsection{Environment Rules: ALFWorld (Tier 3)}

\begin{lstlisting}[caption={Selected environment rules (ALFWorld)}]
heat:
  syntax: heat <object> with microwave 1
  preconditions: [holding(object), at(microwave)]
  effects: [heated(object), holding(object)]
  CRITICAL: Single-step atomic action. Object
  stays in hand. Do NOT put object in microwave.

cool:
  syntax: cool <object> with fridge 1
  preconditions: [holding(object), at(fridge)]
  effects: [cooled(object), holding(object)]
  CRITICAL: Single-step atomic. Do NOT put object
  in fridge manually.

take:
  syntax: take <object> from <location>
  preconditions: [at(location), not holding,
    container open if closed]
  CRITICAL: Can only hold one object.
\end{lstlisting}

\subsection{Environment Rules: ScienceWorld (Tier 3)}

\begin{lstlisting}[caption={Selected environment rules (ScienceWorld)}]
focus:
  preconditions: [target_visible, identity_verified]
  effects: [task_object_identified]
  CRITICAL: wrong object -> immediate score=-100
    and termination; most dangerous action.

thermometer:
  syntax: use thermometer in inventory on <target>
  CRITICAL: stays in hand after measurement.

mix:
  preconditions: [substances combined in container]
  CRITICAL: hand becomes empty after mix.
\end{lstlisting}

\FloatBarrier
\section{Episodic Memory Entry Examples}
\label{app:memory}

The ALFWorld memory store contains three entry types. After Round~1 learning, the store holds 64 success snippets, 157 failure lessons, and 12 verified schemas (233 total). ScienceWorld adds a fourth type: critical failures.

\subsection{Success Snippet (ALFWorld)}

\begin{lstlisting}[caption={Success snippet: sub-goal ``transformed'' (heat task)}]
{
  "id": "success_000008",
  "type": "success",
  "goal_signature":
    "pick_heat_then_place_in_recep:mug:hot:cabinet",
  "state_signature": "holding=none",
  "sub_goal_type": "transformed",
  "precondition": {
    "holding": null,
    "action_type": "heat",
    "tool_location": "microwave 1"
  },
  "actions": [
    "open fridge 1",
    "go to countertop 1",
    "take mug 2 from countertop 1",
    "go to microwave 1",
    "open microwave 1",
    "heat mug 2 with microwave 1"
  ],
  "outcome": {"heat": "mug 2"},
  "success_count": 1
}
\end{lstlisting}

\subsection{Failure Lesson (ALFWorld)}

\begin{lstlisting}[caption={Failure lesson: precondition violation (ALFWorld)}]
{
  "id": "failure_000090",
  "type": "failure",
  "goal_signature":
    "pick_heat_then_place_in_recep:potato:hot:fridge",
  "state_signature":
    "holding=none|open=fridge 1,microwave 1",
  "failed_action":
    "take potato 1 from countertop 3",
  "failure_type": "precondition_not_met",
  "failure_message": "Nothing happens.",
  "corrective_rule":
    "must be at location and container must be
     open before taking",
  "occurrence_count": 3
}
\end{lstlisting}

\subsection{Verified Schema (ALFWorld)}

\begin{lstlisting}[caption={Verified schema: cool action (ALFWorld)}]
{
  "id": "schema_000222",
  "type": "schema",
  "action_template": "cool {object} with fridge 1",
  "action_type": "cool",
  "preconditions": [
    "holding target object",
    "at cool tool location"
  ],
  "effects": [
    "object cooled",
    "still holding object"
  ],
  "success_count": 1,
  "failure_count": 0,
  "confidence": 1.0
}
\end{lstlisting}

\subsection{Critical Failure (ScienceWorld)}

\begin{lstlisting}[caption={Critical failure entry (ScienceWorld)}]
{
  "id": "failure_000078",
  "type": "critical_failure",
  "goal_signature": "F3:melt:water",
  "state_signature":
    "room=kitchen|inventory=empty|progress=0",
  "failed_action": "focus on stove",
  "failure_message": "score = -100, terminated",
  "corrective_rule": "verify object identity
    matches task target before executing focus",
  "occurrence_count": 3,
  "retrieval_weight": 2.0
}
\end{lstlisting}

\FloatBarrier
\section{Statistical Significance}
\label{app:significance}

\paragraph{McNemar's test.}
We report McNemar's test for paired binary outcomes on the 134 unseen tasks (computed on the soft-only RPMS variant, 71/134 = 53.0\%; see Sec.~\ref{sec:arbitration} for arbitration modes).
$p$ denotes the two-sided exact $p$-value; $n_{01}/n_{10}$ counts discordant pairs.

\textbf{Statistically significant results.}
Rules-only vs.\ Baseline ($n_{01}/n_{10}=36/16$, $p=0.008$): rule retrieval produces a robust, independently verifiable gain.
RPMS (soft-only) vs.\ Baseline ($43/20$, $p=0.006$): the full system improvement is similarly significant.
RPMS (soft-only) vs.\ Memory-only ($37/21$, $p=0.049$): the rule-grounded system significantly outperforms unfiltered memory.

\textbf{Expected non-significance.}
RPMS (soft-only) vs.\ Rules-only ($28/25$, $p=0.783$): this comparison is non-significant at the task level, which is consistent with our design intent.
Memory's role in RPMS is not to independently solve additional tasks but to provide applicability-checked experience that amplifies rule effectiveness---a function that is not captured by a pairwise task-success count.
The appropriate test for this claim is the SCF ablation (Appendix~\ref{app:extended_analysis}), which shows that disabling state-consistent filtering costs --6.0~pp even when rules are present, confirming that filtered memory adds value within the rule-grounded system.

\textbf{Note.}
These statistics are computed from the soft-only arbitration variant (71/134 = 53.0\%) because per-task binary outcome labels were collected at that stage of the experimental pipeline.
The hard+soft full system (80/134 = 59.7\%) uses the same frozen memory; its significance over Baseline is expected to be at least as strong, given the larger observed effect size.

\end{document}